\documentclass[10pt, a4paper]{article}
\usepackage{lrec2022} 
\usepackage{multibib}
\newcites{languageresource}{Language Resources}
\usepackage{graphicx}
\usepackage{tabularx}
\usepackage{soul}
\usepackage{comment}
\usepackage{amsmath}
\usepackage{titlesec}
\titleformat{\section}{\normalfont\large\bfseries\center}{\thesection.}{1em}{}
\titleformat{\subsection}{\normalfont\SmallTitleFont\bfseries\raggedright}{\thesubsection.}{1em}{}
\titleformat{\subsubsection}{\normalfont\normalsize\bfseries\raggedright}{\thesubsubsection.}{1em}{}
\renewcommand\thesection{\arabic{section}}
\renewcommand\thesubsection{\thesection.\arabic{subsection}}
\renewcommand\thesubsubsection{\thesubsection.\arabic{subsubsection}}

\usepackage{epstopdf}
\usepackage[utf8]{inputenc}

\usepackage{hyperref}
\usepackage{xstring}

\usepackage{color}

\title{HilMeMe: A Human-in-the-Loop Machine Translation Evaluation Metric \\ Looking into Multi-Word Expressions}

\name{Lifeng Han
} 

\address{ The University of Manchester \\
         lifeng.han@manchester.ac.uk
}

\abstract{
With the fast development of Machine Translation (MT) systems, especially the new boost from Neural MT (NMT) models, the MT output quality has reached a new level of accuracy. However, many researchers criticised that the current popular evaluation metrics such as BLEU can not correctly distinguish the state-of-the-art NMT systems regarding quality differences. In this short paper, we describe the design and implementation of  a linguistically motivated human-in-the-loop evaluation metric looking into idiomatic and terminological Multi-word Expressions (MWEs). MWEs have played a bottleneck in many Natural Language Processing (NLP) tasks including MT. MWEs can be used as one of the main factors to distinguish different MT systems by looking into their capabilities on recognising and translating MWEs in an accurate and meaning equivalent manner.
 \\ \newline  }

\begin{document}

\maketitleabstract

\section{Introduction}
\label{section_intro}
Machine Translation Evaluation (MTE) has been a long-term challenging research topic since the development of MT. MTE plays an important role in MT development and quality evaluation. 
Popular automatic evaluation metrics have failed to correctly distinguish Neural MT (NMT) systems and to distinguish them from the real human parity in very recent expert-based MT evaluation validations \cite{google2021human_evaluation_TQA,L_ubli_2020_human_parity,HanJonesSmeatonBolzoni2021decomposition4mt_MWE,han2022investigation}. With this thought in mind, human evaluation is still very much needed to correctly indicate the progress of current state-of-the-art MT systems. 
On the one hand, most frequently used human evaluation methods still focus on sentence level accuracy of translation outputs, ignoring domain-specific terminology and linguistic phenomena, such as metaphorical phrases, idiomatic Multi-word Expressions (MWEs), information weights on different words and particles within a sentence. Such human evaluation methods include 
human-targeted translation edit rate (HTER) \cite{SnoverDorrSchwartzMicciulla2006} and Direct Assessment (DA) \cite{graham_baldwin_moffat_zobel_2017}. 
On the other hand, the open-sourced Multi-dimension Quality (MQM) metric \cite{MQM2014} has been developed into tremendous details which is too time-consuming to apply and also requires a very skilled evaluator. 
With the development and availability of MWE-annotated corpora, it becomes capable to incorporate MWEs as part of the evaluation metric component. Such corpora with MWE annotations include monolingual ones, such as PARSEME shared task data on MWE identification and discovery \cite{PARSEME2018task,parseme2017shared}, as well as multilingual parallel ones such as AlphaMWE \cite{han-etal-2020-alphamwe} which includes Chinese, German, Polish, and Italian all being translated and aligned from English root corpus \cite{mwe2018english}.
The parallel annotation of AlphaMWE corpus allows us to stay within sentence-level rubric grading for the overall translation accuracy, but also going sufficiently deeper into the domain-level linguistic aspects.

In this work, we address the current MTE issue by designing a linguistically motivated and human-in-the-loop MTE metric looking into MWEs (HilMeMe). For domain-specific content, the MWE annotated corpus can be prepared on the fly in the course of translation evaluation process.
MWEs have been a big challenge for many Natural Language Understanding (NLU) and Natural Language Processing (NLP) tasks due to their idiomaticity, low-frequency, and richness in variety. 
The study on MWEs involves a broad list of research topics, such as idioms, metaphors, slang, fixed and semi-fixed expressions, and compound words, etc. \cite{Sag2002MWE,mwe2017survey}. 

In this paper, we describe the methodological design of HilMeMe and its implementation. We will make the platform open-source for research purpose at \url{https://github.com/poethan/HilMeMe}. 
This methodology takes MWEs as one important factor in the assessment procedure, in addition to judging the overall contextual sentence or segment level translation performances. 
It asks  assessors to do certain level classification of the error types regarding MWE translations, e.g. if it is translated correctly or not, using reference MWEs or alternatives or common phrases, etc. These classification behaviours are saved in our toolkit and can be exported for researchers to carry out further analysis of their system outputs. With this in mind, we explore whether HiLMeMe can have a positive influence for MT modelling research, as well as during actual production of machine-edited translations.

The rest of the paper is organised as below: Section \ref{section_related} presents the related work to ours, Section \ref{section_method} and Section \ref{section_implement} introduce the methodological design of HilMeMe and the implementation of the platform, and Section \ref{section_conclu} concludes this paper with future work.

\section{Related Work}
\label{section_related}
The earliest human assessment methods for MT can be traced back to around 1966. They include the intelligibility and fidelity used by the automatic language processing advisory committee (ALPAC) \cite{Carroll1966}.

The requirement that a translation is intelligible means that, as far as possible, the translation should read like normal, well-edited prose, and be readily understandable in the same way that such a translation would be understandable if originally written by a native speaker of the target language (and not as translation). The requirement that a translation is of high fidelity or accuracy includes the requirement that the translation should, as little as possible, twist, distort, or controvert the meaning intended by the original.

In the 1990s, the Advanced Research Projects Agency (ARPA) created a methodology to evaluate MT systems using adequacy, fluency and comprehension of MT output \cite{ChurchHovy1991}, which was subsequently adapted for use in MT evaluation campaigns including \cite{WhiteConnellMara1994}.
To set up this methodology, a human assessor is asked to look at each fragment, delimited by syntactic constituents and containing sufficient information, and judge its adequacy on a scale of 1-to-5. Results are computed by averaging the judgements over all of the decisions in the translation set.

Fluency evaluation is compiled in the same manner as for adequacy except that the assessor is asked to make intuitive judgements on a sentence-by-sentence basis for each translation. Human assessors are asked to determine whether the translation is good English without reference to the correct translation. Fluency evaluation determines whether a sentence is well-formed and fluent in context.

Comprehension relates to ``Informativeness'', whose objective is to measure a system's ability to produce a translation that conveys sufficient information, such that people can gain necessary information from it. The reference set of expert translations is used to create six questions with six possible answers respectively including, ``none of the above'' and ``cannot be determined''.

Work by White and Taylor \cite{WhiteTaylor1998} developed a task-oriented evaluation methodology for Japanese-to-English translation to measure MT systems in light of the tasks for which their output might be used. They seek to associate the diagnostic scores assigned to the output used in the DARPA (Defense Advanced Research Projects Agency) evaluation with a scale of language-dependent tasks, such as scanning, sorting, and topic identification.
Another tasked-based MT output evaluation by the extraction of three types of elements namely: \textit{who}, \textit{when}, and  \textit{where} was, introduced in \cite{Voss06task-basedevaluation}.

One example of a metric that is designed using post-editing is HTER \cite{SnoverDorrSchwartzMicciulla2006} which is based on the translation edit rate (TER) metric \cite{Olive-2005-TER} using the number of editing steps. 
Here, a human assessor has to find the minimum number of insertions, deletions, substitutions, and shifts to convert the system output into an acceptable translation. 
HTER calculates the minimum of edits to a \textit{new targeted reference}, i.e. the post-edited translation. %

\newcite{graham-etal-2013-continuous} noted that the lower agreements from WMT human assessment might be caused partially by the interval-level scales set up for the human assessor to make a quality judgement of each segment. For instance, the human assessor 
might be in a situation where neither of the two categories they were forced to choose is preferred. In light of this rationale, they proposed continuous measurement scales (CMS) for human translation quality assessment (TQA) using fluency criteria. This was implemented by introducing the crowd-sourcing platform Amazon Mechanical Turk (MTurk), which has been popular in both NLP and multimedia research tasks \cite{graham_baldwin_moffat_zobel_2017,graham-etal-2020-power_translationese}.
However, the problem with Mechanical Turk is that the evaluators are not trained linguists and lack of domain specific knowledge. The unskilled translators have two major problems: 1) they tend to rate literal translations with higher quality than actuarial good translations which inflates translation quality score, and 2) they tend not to see domain-specific errors due to the lack of domain knowledge and terminology, or even understanding of the meaning.

Our recent work ``HOPE'' was introduced as an attempt to address part of these
 issues by using professional human translators as coders and using some refined error categories for the coders to look into and score the error accordingly \cite{gladkoff-han-2022-hope}. In addition, we introduced error severity levels for each of the error category. 

HilMeMe model takes full advantage of the experiences of aforementioned methods such as setting up the sentence-level fluency and adequacy factors that have been proved useful, avoiding translation pair comparison and taking direct scoring as preferred option that has been tested more effective, and designing specific features from task-oriented paradigm. However, none of above work has focused on MWEs as linguistic component in their evaluation method design. We refer to \citelanguageresource{Mapelli2019elra} and \cite{han-etal-2020-alphamwe} for examples on the difficulty of idiomatic MWE translations, as well as domain-specific knowledge component.

\section{Methodological Design}
\label{section_method}

\begin{figure*}[!th]
\begin{center}
\centering
\includegraphics*[width=\textwidth]{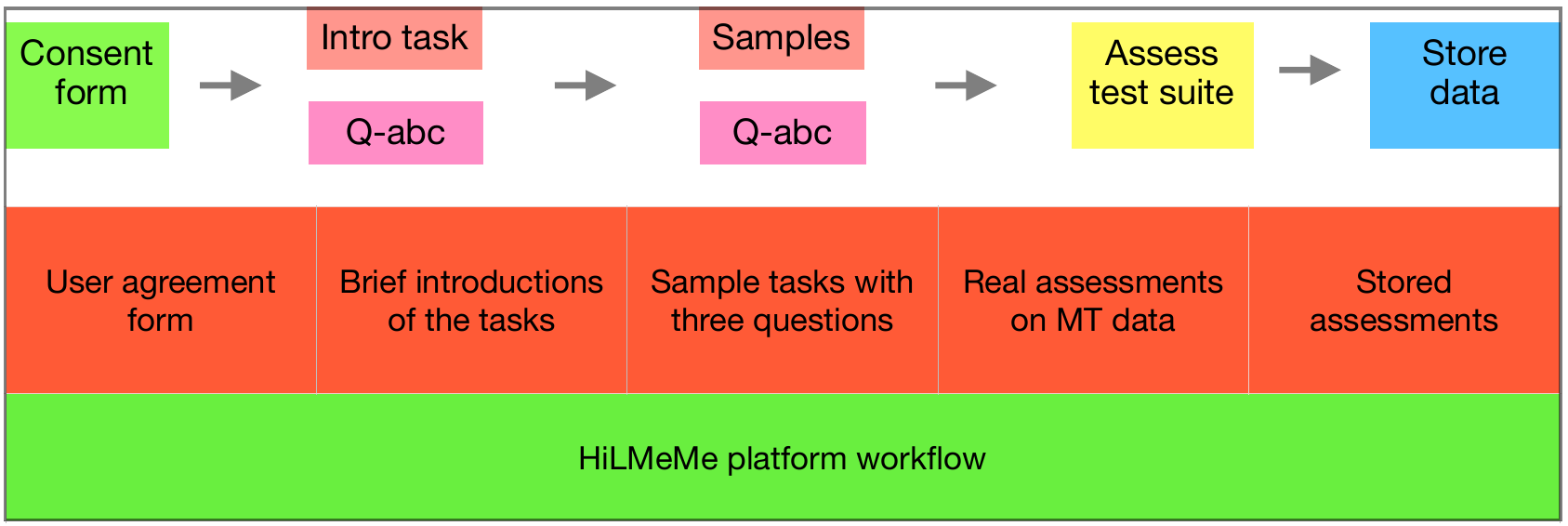}
\caption{HiLMeMe Platform Workflow }
\label{fig:hilmeme_workflow}
\end{center}
\end{figure*}

\begin{figure*}[!h]
\begin{center}
\centering
\includegraphics*[width=\textwidth]{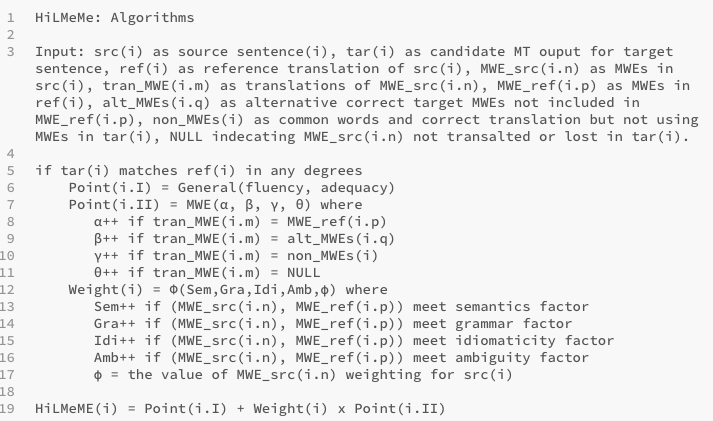}
\caption{HiLMeMe Algorithms}
\label{fig:hilmeme_pseudo}
\end{center}
\end{figure*}

HiLMeMe can be placed into the area of semantic features integrated human assessment, with domain-specific and knowledge-based MWEs as the lexical terms featured, in addition to fluency and adequacy criteria being used. It is also connected to task-oriented evaluations.
There is a  three-step assessment task designed in HiLMeMe, including general text (sentences, segments) level fluency and adequacy score $General(fluency, adequacy)$, highlighted MWEs translation quality score $MWE(\alpha, \beta, \gamma, \theta)$, and a weighting parameter for MWEs on overall text $\Phi$. We describe these separately below before we get to the computation of the overall HiLMeMe score (see the pseudo algorithms in Figure \ref{fig:hilmeme_pseudo} 
).

\noindent I. How good is the MT output text in general?
\begin{itemize}
     \item Look at the two factors when scoring.
     \item Fluency: Is the candidate translation fluent, e.g. grammatically correct
     \item Adequacy: Does the candidate translation cover all the meaning in the source / reference text?
     \item Give a score 0 to 10   $\rightarrow General(fluency, adequacy)$.
\end{itemize}

In this interface, we give a scoring range 0 to 10.

\noindent II. Look into the highlighted MWEs and classify if they are translated, and if so then how?
\begin{itemize}
     \item Correctly translated using reference MWEs ($\alpha++$, score:10)
     \item Correctly translated using alternative MWEs ($\beta++$, score:10)
     \item Translated using other words, non-MWEs ($\gamma++$, score:0 to 10)
     \item Not translated, lost, NULL ($\theta$, score:0)
     \item Score $\rightarrow MWE(\alpha, \beta, \gamma, \theta)$
\end{itemize}

In this interface, we give four choices \textit{ref-MWE}, \textit{alt-MWE}, \textit{non-MWE}, and \textit{NULL}, in addition to a scoring range 0 to 10. The triple set ($\alpha$, $\beta$, $\gamma$ ) stores how often the MWEs are translated using the reference MWE, alternative MWE, or other words, and $\theta$ stores how often the source MWEs are left in a loss in the translation or kept as foreign words without any translation.

\noindent III. From which aspects do the MWEs present difficulty, affect the translation, and to what degrees?

\begin{itemize}
     \item Semantics: word meanings and relations between them
     \item Grammar: syntax and morphology
     \item Idiomaticity: a group of words established by usage as having a meaning not deducible from those of the individual words
     \item Ambiguity: the quality of being open to more than one interpretation, including domain-specific terms, expressions and language
     \item Degrees ($\phi$, score:0 to 1)
     \item Output $\rightarrow \Phi(Sem, Gra, Idi, Amb, \phi)$
\end{itemize}

\noindent where the parameters $Sem, Gra, Idi, Amb$ represent \textit{semantics}, \textit{grammar}, \textit{idiomaticity} and \textit{ambiguity} respectively. The classification of different situations in  step III is to facilitate further analysis on the MWEs appearing in our test set (corpus), as well as the possible extension in  future to cover more labelled data with broader aspects. This is a multiple choice classification where the assessors can tick more than one of the categories. 

Finally, the overall score of HiLMeMe, i.e. $HiLMeMe(General, MWE, \Phi)$, is the weighted sum of the general text score  $General(fluency, adequacy)$ and MWE score $MWE(\alpha, \beta, \gamma, \theta)$ with the weighting parameter from the third step $\phi$ on the influence of MWEs on overall text.

The scoring function is as below and is based on the three step judgements where we use $HiLMeMe(\bullet)$ to indicate $HiLMeMe(General, MWE, \Phi)$.


\begin{gather*} 
    HiLMeMe(\bullet) = General(fluency, adequacy)+\\
    \phi \times MWE(\alpha, \beta, \gamma, \theta)
 \end{gather*}

\begin{gather*} 
    HiLMeMe_{norm}= HiLMeMe/Point_{Max}
 \end{gather*}

\noindent where $Point_{Max}$ is the maximum point that  step-I and step-II can generate, and $HiLMeMe_{norm}$ is the normalised score propagating the HiLMeMe score into the interval (0, 1).

The overall score is the combination of step-I and step-II with a weighting parameter attached to the second point. The normalised score of HiLMeMe is the raw score divided by the highest potential score they can get, such that the normalised score ranges from 0 to 1. The normalisation is to give the user a more straightforward instinct on how much the assessors judge the translation text quality in a 0-to-1 (0 to 100) range. Another benefit of the normalisation is that it can be used  for automatic evaluation metrics tuning by calculating their correlation to HiLMeMe, e.g. Spearman, Pearson, or Kendall Tau correlation methods at system level or segment level \cite{han-etal-2021-TQA}.

This methodology can also be used to create new resources. For example, the step-II MWE question where we ask if the translation uses alternative MWEs and if so then which, here we can set a further storing option to save the alternative MWEs that are correct translations of the source MWEs. We can also store  plain phrases that are correct translations of source MWEs. In this way, we generate more bilingual parallel MWE terms, including paraphrasing at single side at MWE level. 
These resources can be important linguistic driven knowledge base features for popular automatic evaluation metrics such as METEOR \cite{BanerjeeLavie2005,DBLP:conf/coling/ServanBEBB16}, which depends on high quality paraphrase data  to achieve better evaluations. 
From the translation modelling perspective, the  extracted and stored multilingual paraphrased MWEs can be integrated into MT modelling  learning and translating to generate alternative high quality translation outputs with lexical diversity.
Furthermore, paraphrase databases are widely used in NLP communities in different tasks such as natural language inference, natural language understanding, text entailment, searching, etc.

\section{HiLMeMe Implementation and the Platform}
\label{section_implement}

HiLMeMe is implemented via the PsychoPy3 platform, relying on Python3 packages. 
PsychoPy\footnote{ \url{https://www.psychopy.org}} has been a popular platform for researchers to carry out experiments, especially in the situations where human interactive or assessments are needed. These behavioural sciences include neuroscience, psychology, psychophysics, and linguistics. It can easily accommodate our human-in-the-loop evaluation methodology by offering a straightforward interface and storing all the classification data during the assessments. 
The HiLMeMe initial PsychoPy3 platform is available and  will be open source and publicly available.
In the implemented platform, we designed the following HiLMeMe workflow as shown in Figure~\ref{fig:hilmeme_workflow}: 
with the following sequence of steps: consent form $\rightarrow$ task introduction $\rightarrow$ sample practice with three questions $\rightarrow$ real assessment of MT results $\rightarrow$ stored assessment data.

\section{Conclusion and Future Work}
\label{section_conclu}

To achieve more reliable MT quality assessment and advance the state-of-the-art in MT modelling, we designed a new evaluation methodology, having a human-in-the-loop and looking specifically into MWEs.
We introduced three-step assessment models with corresponding questions and error classifications. We presented the scoring functions based on the three main questions and the implemented platform. We discussed the potential impact of HiLMeMe and its output data.

HiLMeMe is based on multilingual parallel corpus with MWE annotations such as AlphaMWE \cite{han-etal-2020-alphamwe} (data page \footnote{https://github.com/poethan/AlphaMWE}), as an example and the model for domain-specific annotated expressions and terminology.
We expect this new evaluation method to reflect the differences of state-of-the-art MT models in performance \textit{towards human parity}, e.g. translation in a situation with idiomatic and ambiguous phrases, knowledge-specific and domain-specific expressions. 
We will open source our platform to MT researchers and the NLP evaluation community, as well as translation and localisation industry practitioners, 
for better evaluating their MT models, and for better correlating their evaluation metrics with expert evaluations. In the future, we will also release our expert scored data using HilMeMe on different MT systems from deployed language pairs that are under development. 

\section*{Acknowledgement}
The authors thank Prof Alan Smeaton and Prof Gareth Jones for valuable advice on this work, thank Sonia Ramotowska for the instruction of PsychoPy platform during the initial HiLMeMe platform implementation, thank Gabriella Guagliardo on the name endorsement of HiLMeMe, and thank Serge Gladkoff for manuscript wording revision.
This work was partially funded by ADAPT Research Centre, DCU, Ireland.

\section{Bibliographical References}\label{reference}

\bibliographystyle{lrec2022-bib}
\bibliography{lrec2022-example}

\begin{thebibliography}{}

\bibitem[\protect\citename{{Valérie Mapelli }}2019]{Mapelli2019elra}
{Valérie Mapelli }.
\newblock (2019).
\newblock {\em Chinese-English Database of Proverbs and Idioms (Chengyu)
  Lexical Conceptual}.
\newblock ELRA, ISLRN 506-728-933-717-0.

\end{thebibliography}


\begin{thebibliography}{}

\bibitem[\protect\citename{Banerjee and Lavie}2005]{BanerjeeLavie2005}
Banerjee, S. and Lavie, A.
\newblock (2005).
\newblock Meteor: An automatic metric for mt evaluation with improved
  correlation with human judgments.
\newblock In {\em Proceedings of the ACL}.

\bibitem[\protect\citename{Carroll}1966]{Carroll1966}
Carroll, J.~B.
\newblock (1966).
\newblock An experiment in evaluating the quality of translation.
\newblock {\em Mechanical Translation and Computational Linguistics},
  9(3-4):67--75.

\bibitem[\protect\citename{Church and Hovy}1991]{ChurchHovy1991}
Church, K. and Hovy, E.
\newblock (1991).
\newblock Good applications for crummy machine translation.
\newblock In {\em Proceedings of the Natural Language Processing Systems
  Evaluation Workshop}.

\bibitem[\protect\citename{Constant \bgroup et al.\egroup }2017]{mwe2017survey}
Constant, M., Eryi{\v{g}}it, G., Monti, J., van~der Plas, L., Ramisch, C.,
  Rosner, M., and Todirascu, A.
\newblock (2017).
\newblock {S}urvey: Multiword expression processing: A {S}urvey.
\newblock {\em Computational Linguistics}, 43(4):837--892.

\bibitem[\protect\citename{{Freitag} \bgroup et al.\egroup
  }2021]{google2021human_evaluation_TQA}
{Freitag}, M., {Foster}, G., {Grangier}, D., {Ratnakar}, V., {Tan}, Q., and
  {Macherey}, W.
\newblock (2021).
\newblock {Experts, Errors, and Context: A Large-Scale Study of Human
  Evaluation for Machine Translation}.
\newblock {\em arXiv e-prints}, page arXiv:2104.14478, April.

\bibitem[\protect\citename{Gladkoff and Han}2022]{gladkoff-han-2022-hope}
Gladkoff, S. and Han, L.
\newblock (2022).
\newblock {HOPE}: A task-oriented and human-centric evaluation framework using
  professional post-editing towards more effective {MT} evaluation.
\newblock In {\em Proceedings of the Thirteenth Language Resources and
  Evaluation Conference}, pages 13--21, Marseille, France, June. European
  Language Resources Association.

\bibitem[\protect\citename{Graham \bgroup et al.\egroup
  }2013]{graham-etal-2013-continuous}
Graham, Y., Baldwin, T., Moffat, A., and Zobel, J.
\newblock (2013).
\newblock Continuous measurement scales in human evaluation of machine
  translation.
\newblock In {\em Proceedings of the 7th Linguistic Annotation Workshop and
  Interoperability with Discourse}, pages 33--41, Sofia, Bulgaria, August.
  Association for Computational Linguistics.

\bibitem[\protect\citename{Graham \bgroup et al.\egroup
  }2017]{graham_baldwin_moffat_zobel_2017}
Graham, Y., Baldwin, T., Moffat, A., and Zobel, J.
\newblock (2017).
\newblock Can machine translation systems be evaluated by the crowd alone.
\newblock {\em Natural Language Engineering}, 23(1):3–30.

\bibitem[\protect\citename{Graham \bgroup et al.\egroup
  }2020]{graham-etal-2020-power_translationese}
Graham, Y., Haddow, B., and Koehn, P.
\newblock (2020).
\newblock Statistical power and translationese in machine translation
  evaluation.
\newblock In {\em Proceedings of the 2020 Conference on Empirical Methods in
  Natural Language Processing (EMNLP)}, pages 72--81, Online, November.
  Association for Computational Linguistics.

\bibitem[\protect\citename{Han \bgroup et al.\egroup
  }2020]{han-etal-2020-alphamwe}
Han, L., Jones, G., and Smeaton, A.
\newblock (2020).
\newblock {A}lpha{MWE}: Construction of multilingual parallel corpora with
  {MWE} annotations.
\newblock In {\em Proceedings of the Joint Workshop on Multiword Expressions
  and Electronic Lexicons}, pages 44--57, online, December. Association for
  Computational Linguistics.

\bibitem[\protect\citename{Han \bgroup et al.\egroup
  }2021a]{HanJonesSmeatonBolzoni2021decomposition4mt_MWE}
Han, L., Jones, G., Smeaton, A., and Bolzoni, P.
\newblock (2021a).
\newblock {C}hinese character decomposition for neural {MT} with multi-word
  expressions.
\newblock In {\em Proceedings of the 23rd Nordic Conference on Computational
  Linguistics (NoDaLiDa)}, pages 336--344, Reykjavik, Iceland (Online), May
  31--2 June. Link{\"o}ping University Electronic Press, Sweden.

\bibitem[\protect\citename{Han \bgroup et al.\egroup }2021b]{han-etal-2021-TQA}
Han, L., Smeaton, A., and Jones, G.
\newblock (2021b).
\newblock Translation quality assessment: A brief survey on manual and
  automatic methods.
\newblock In {\em Proceedings for the First Workshop on Modelling Translation:
  Translatology in the Digital Age}, pages 15--33, online, May. Association for
  Computational Linguistics.

\bibitem[\protect\citename{Han}2022]{han2022investigation}
Han, L.
\newblock (2022).
\newblock {\em An investigation into multi-word expressions in machine
  translation}.
\newblock {Ph.D.} thesis, Dublin City University.

\bibitem[\protect\citename{Lommel \bgroup et al.\egroup }2014]{MQM2014}
Lommel, A., Burchardt, A., and Uszkoreit, H.
\newblock (2014).
\newblock Multidimensional quality metrics (mqm): A framework for declaring and
  describing translation quality metrics.
\newblock {\em Tradumàtica: tecnologies de la traducció}, 0:455--463, 12.

\bibitem[\protect\citename{Läubli \bgroup et al.\egroup
  }2020]{L_ubli_2020_human_parity}
Läubli, S., Castilho, S., Neubig, G., Sennrich, R., Shen, Q., and Toral, A.
\newblock (2020).
\newblock A set of recommendations for assessing human–machine parity in
  language translation.
\newblock {\em Journal of Artificial Intelligence Research}, 67, Mar.

\bibitem[\protect\citename{Olive}2005]{Olive-2005-TER}
Olive, J.
\newblock (2005).
\newblock Global autonomous language exploitation (gale).
\newblock In {\em DARPA/IPTO Proposer Information Pamphlet}.

\bibitem[\protect\citename{Ramisch \bgroup et al.\egroup
  }2018]{PARSEME2018task}
Ramisch, C., Cordeiro, S.~R., Savary, A., Vincze, V., Barbu~Mititelu, V.,
  Bhatia, A., Buljan, M., Candito, M., Gantar, P., Giouli, V., G{\"u}ng{\"o}r,
  T., Hawwari, A., I{\~n}urrieta, U., Kovalevskait{\.e}, J., Krek, S., Lichte,
  T., Liebeskind, C., Monti, J., Parra~Escart{\'\i}n, C., QasemiZadeh, B.,
  Ramisch, R., Schneider, N., Stoyanova, I., Vaidya, A., and Walsh, A.
\newblock (2018).
\newblock Edition 1.1 of the {PARSEME} shared task on automatic identification
  of verbal multiword expressions.
\newblock In {\em Proceedings of the Joint Workshop on Linguistic Annotation,
  Multiword Expressions and Constructions ({LAW}-{MWE}-{C}x{G}-2018)}, pages
  222--240, Santa Fe, New Mexico, USA, August. Association for Computational
  Linguistics.

\bibitem[\protect\citename{Sag \bgroup et al.\egroup }2002]{Sag2002MWE}
Sag, I.~A., Baldwin, T., Bond, F., Copestake, A., and Flickinger, D.
\newblock (2002).
\newblock Multiword expressions: A pain in the neck for nlp.
\newblock In Alexander Gelbukh, editor, {\em Computational Linguistics and
  Intelligent Text Processing}, pages 1--15, Berlin, Heidelberg. Springer
  Berlin Heidelberg.

\bibitem[\protect\citename{Savary \bgroup et al.\egroup
  }2017]{parseme2017shared}
Savary, A., Ramisch, C., Cordeiro, S., Sangati, F., Vincze, V., QasemiZadeh,
  B., Candito, M., Cap, F., Giouli, V., Stoyanova, I., and Doucet, A.
\newblock (2017).
\newblock The {PARSEME} shared task on automatic identification of verbal
  multiword expressions.
\newblock In {\em Proceedings of the 13th Workshop on Multiword Expressions
  ({MWE} 2017)}, pages 31--47, Valencia, Spain.

\bibitem[\protect\citename{Servan \bgroup et al.\egroup
  }2016]{DBLP:conf/coling/ServanBEBB16}
Servan, C., Berard, A., Elloumi, Z., Blanchon, H., and Besacier, L.
\newblock (2016).
\newblock Word2vec vs dbnary: Augmenting {METEOR} using vector representations
  or lexical resources?
\newblock In {\em {COLING} 2016, 26th International Conference on Computational
  Linguistics, Proceedings of the Conference: Technical Papers, December 11-16,
  2016, Osaka, Japan}, pages 1159--1168.

\bibitem[\protect\citename{Snover \bgroup et al.\egroup
  }2006]{SnoverDorrSchwartzMicciulla2006}
Snover, M., Dorr, B.~J., Schwartz, R., Micciulla, L., and Makhoul, J.
\newblock (2006).
\newblock A study of translation edit rate with targeted human annotation.
\newblock In {\em Proceeding of AMTA}.

\bibitem[\protect\citename{Voss and Tate}2006]{Voss06task-basedevaluation}
Voss, C.~R. and Tate, R.~R.
\newblock (2006).
\newblock Task-based evaluation of machine translation (mt) engines: Measuring
  how well people extract who, when, where-type elements in mt output.
\newblock In {\em In Proceedings of 11th Annual Conference of the European
  Association for Machine Translation (EAMT-2006}, pages 203--212.

\bibitem[\protect\citename{Walsh \bgroup et al.\egroup }2018]{mwe2018english}
Walsh, A., Bonial, C., Geeraert, K., McCrae, J.~P., Schneider, N., and Somers,
  C.
\newblock (2018).
\newblock Constructing an annotated corpus of verbal {MWE}s for {E}nglish.
\newblock In {\em Proceedings of the Joint Workshop on Linguistic Annotation,
  Multiword Expressions and Constructions ({LAW}-{MWE}-{C}x{G}-2018)}, pages
  193--200, Santa Fe, New Mexico, USA, August. Association for Computational
  Linguistics.

\bibitem[\protect\citename{White and Taylor}1998]{WhiteTaylor1998}
White, J.~S. and Taylor, K.~B.
\newblock (1998).
\newblock A task-oriented evaluation metric for machine translation.
\newblock In {\em Proceeding LREC}.

\bibitem[\protect\citename{White \bgroup et al.\egroup
  }1994]{WhiteConnellMara1994}
White, J.~S., Connell, T.~O., and Mara, F.~O.
\newblock (1994).
\newblock The arpa mt evaluation methodologies: Evolution, lessons, and future
  approaches.
\newblock In {\em Proceeding of AMTA}.

\end{thebibliography}

\section{Language Resource References}
\label{lr:ref}
\bibliographystylelanguageresource{lrec2022-bib}
\bibliographylanguageresource{languageresource}


\end{document}